\documentclass{sigchi}

\CopyrightYear{2016}
\setcopyright{acmlicensed}
\doi{http://dx.doi.org/10.475/123_4}
\isbn{123-4567-24-567/08/06}
\conferenceinfo{CHI'16,}{May 07--12, 2016, San Jose, CA, USA}
\acmPrice{\$15.00}

\toappear{Submission to IUI 2019. Submission \#: 1166.}

\usepackage{balance}       
\usepackage{graphics}      
\usepackage[T1]{fontenc}   
\usepackage{txfonts}
\usepackage{mathptmx}
\usepackage[pdflang={en-US},pdftex]{hyperref}
\usepackage{color}
\usepackage{booktabs}
\usepackage{textcomp}
\usepackage{hanging}
\usepackage{libertine}
\usepackage{amsmath}
\usepackage{tabu}
\usepackage{adjustbox}

\usepackage{microtype}        
\usepackage{ccicons}          

\usepackage{todonotes}

\definecolor{mygray}{gray}{0.6}
\newcommand{\name}{Challenge.AI} 
\newcommand{\etal}{et~al.}
\newcommand{\eg}{e.g.}
\newcommand{\ie}{i.e.}

\newcommand{\siwei}[1]{\textcolor{black}{#1}}
\newcommand{\fade}[1]{\textcolor{mygray}{#1}}

\expandafter\def\expandafter\normalsize\expandafter{%
    \normalsize
    \setlength\abovedisplayskip{8pt}
    \setlength\belowdisplayskip{8pt}
    \setlength\abovedisplayshortskip{8pt}
    \setlength\belowdisplayshortskip{8pt}
}

\allowdisplaybreaks
\def\plaintitle{Challenge AI's Mind: \\A Crowd System for Proactive AI Testing }
\def\plainauthor{Siwei Fu, Anbang Xu, Xiaotong Liu,
  Hellen Zhou, Rama Akkiraju}

\def\plainkeywords{Crowdsourcing, AI, Proactive testing}

\makeatletter
\def\url@leostyle{%
  \@ifundefined{selectfont}{
    \def\UrlFont{\sf}
  }{
    \def\UrlFont{\small\bf\ttfamily}
  }}
\makeatother
\def\pprw{8.5in}
\def\pprh{11in}

\definecolor{linkColor}{RGB}{6,125,233}
\hypersetup{%
  pdftitle={\plaintitle},
 pdfauthor={\plainauthor},
  pdfkeywords={\plainkeywords},
  pdfdisplaydoctitle=true, 
  bookmarksnumbered,
  pdfstartview={FitH},
  colorlinks,
  citecolor=black,
  filecolor=black,
  linkcolor=black,
  urlcolor=linkColor,
  breaklinks=true,
  hypertexnames=false
}

\begin{document}

\title{\plaintitle}

 
\numberofauthors{1}
\author{%
  \centering
  \alignauthor{Siwei Fu$^{1}$ \quad Anbang Xu$^2$ \quad Xiaotong Liu$^2$ \quad Huimin Zhou$^2$ \quad Rama Akkiraju$^2$\\
    \affaddr{$^1$Hong Kong University of Science and Technology \quad $^2$IBM Research --- Almaden}\\
    \email{ sfuaa@cse.ust.hk \quad \{anbangxu, akkiraju\}@us.ibm.com \quad \{xiaotong.liu, Huimin.Zhou1\}@ibm.com}
  }
}

\maketitle

\begin{abstract}
  Artificial Intelligence (AI) has burrowed into our lives in various aspects; however, without appropriate testing, deployed AI systems are often being criticized to fail in critical and embarrassing cases. Existing testing approaches mainly depend on fixed and pre-defined datasets, providing a limited testing coverage. In this paper, we propose the concept of proactive testing to dynamically generate testing data and evaluate the performance of AI systems. We further introduce \name{}, a new crowd system that features the integration of crowdsourcing and machine learning techniques in the process of error generation, error validation, error categorization, and error analysis. 
  We present experiences and insights into a participatory design with AI developers. 
  The evaluation shows that the crowd workflow is more effective with the help of machine learning techniques. 
  AI developers found that our system can help them discover unknown errors made by the AI models, and engage in the process of proactive testing.
\end{abstract}



\section{Introduction}
Artificial Intelligence (AI) becomes a technology renaissance and is beginning to solve problems in many domains. It often performs well under single-score metrics such as precision and recall. For instance, the state-of-the-art AI model has reduced the test error rates of sentiment classification to 4.6\% \cite{Howard2018}. The AI developments in speech recognition, autonomous vehicle and the mastery of the "Go" 
reveal how much AI is capable of.

Yet, with all of the AI success, many AI applications are also criticized as they can fail in critical and embarrassing cases. Recent AI-powered facial recognition systems of Microsoft, IBM, and Face++ have 34\%  more errors with dark-skinned females than light-skinned males \cite{tay}. Fatal Tesla crash raises serious questions about how to test AI products and ensure their reliability.

To address this problem, we propose \textit{proactive testing}, a novel  approach that evaluates the performance of AI models with dynamic and well-crafted dataset collected using crowd intelligence.  
Proactive testing differs from conventional testing metrics in two aspects. 
First, it extends the coverage of the testing dataset by dynamically collecting external dataset.
Second, AI developers are allowed to query additional dataset belonging to certain categories to target corner cases.
As a result, proactive testing is an approach to discovering unknown error and bias of a model, and providing a comprehensive evaluation of the model's performance regarding all test cases.



However, proactive testing is non-trivial in practice. 
First, it is challenging to ensure the quality of newly crafted dataset to support the testing process and obtain reliable error analysis of the AI models. 
Crafting adversarial samples in text domains has attracted attention in machine learning communities in recent years~\cite{Papernot2016,Gao2018,Ribeiro2018,Ebrahimi2018}, but the machine-generated textural errors can be unnatural and lose semantic meaning compared with natural languages. 
Second, the new dataset should address specific requirements raised by AI developers to enable task-dependent analysis of the AI models. 
Proactive testing should be an iterative and dynamic process that continuously brings in domain knowledge of AI developers to improve the test coverage as they build and test the models.

In this paper, we contribute a hybrid system, \name{}, that combines human intelligence and machine learning techniques to assist AI developers in the process of proactive testing. 
Our system contains four main components: explanation-based error generation, error validation, categorization, and analysis. We bring in crowd force in error generation and encourage the crowd to craft sentences that can fail a given AI model. 
Especially, to assist the crowd in error generation, we borrow advanced machine learning methods to explain each prediction made by the model, and present the explanation to the crowd using intuitive visualization.
In addition, we employ the crowd in error validation and categorization to ensure the quality of the crafted dataset at scale.




We evaluate the effectiveness of explanation-based error generation by measuring the performance of the crowd. The evaluation shows that the crowd spent less time in generating specific errors. Moreover, we evaluate the usefulness of the \name{} system through two rounds of exploratory studies with AI developers. \siwei{We found that our system can help AI developers discover unknown errors made by their AI models.}

To summarize, the contributions of our paper are threefold.
First, we propose \name{}, a crowd system that supports proactive testing for AI models by  extending the coverage of testing dataset with crowd-generated samples.

Second, to assist the crowd craft errors to challenge AI models, we propose an explanation-based error generation technique combining human intelligence and machine learning.

Third, we evaluate the effectiveness and usefulness of \name{} by two rounds of interview sessions with five AI developers. In addition, we use crowd evaluation to compare the explanation-based error generation technique and a baseline approach.

\section{Related Work}
This paper is related to prior work in three areas, \eg,
generation of adversarial samples using machine learning, acquisition of corpus using crowd intelligence, and the effects of various prompts.

\subsection{Adversarial learning for text classifiers}
A quantity of approaches have been proposed to generate adversarial examples in the deep learning community.
However, most studies have focused on attacking image or audio classification models~\cite{Goodfellow2014,Kereliuk2015,Kurakin2016,Papernot2016a,Papernot2017}. The attack of text classifiers is under-exploited due to the discrete domains involved in text~\cite{Zhao2017}.

To craft adversarial samples for text classifiers, some work modify the original input. 
For example, 
Liang \etal~\cite{Liang2017} proposed three perturbation strategies, \eg, insertion, deletion, and replacement to evade DNN-based text classifiers. 
Li \etal~\cite{Li2016} studied the effect of removal of input text at different levels of representation.
Gao \etal~\cite{Gao2018} proposed novel scoring strategies to identify critical tokens and executed a modification on those tokens.
Similarly, HotFlip \etal~\cite{Ebrahimi2018} edit the input text at the character level.
Ribeiro \etal \cite{Ribeiro2018} move one step forward by manipulating the input at the word level. That is, replacing tokens by random words of the same POS tag. 
Given access to the model's architecture, \eg, the computational graph, Papernot \etal~\cite{Papernot2016} manipulate the output of RNN models.
Although aforementioned approaches can generate sentences that fail text classifiers, the perturbation harms text integrity, resulting in unnatural and semantically meaningless text from language viewpoint.

To overcome the limitation of above methods, 
Samanta \etal~\cite{Samanta2017} proposed a rule-based approach to ensure that the resulting text is syntactically correct.
Zhao~\etal~\cite{Zhao2017} proposed GAN-based approach to generate adversarial input that are legible to humans.
The two techniques driven by machine learning are promising due to large scalability.
However, the resulting text has not been validated, and its quality is not guaranteed.
In this paper, 
we design a crowdsourcing pipeline to generate and validate adversarial samples by means of human intelligence.
The derived adversarial dataset is diverse from different perspectives.



\subsection{Corpus acquisition using crowdsourcing}
Online crowdsourcing provides easy and economic access to human talent~\cite{Mitchell2014}, and has been proved effective in the acquisition of corpus in various natural language processing tasks. 
Some work focuses on speech transcription. For example,
Parent~\cite{Parent2010} proposed a two-stage approach to transcribe large
amounts of speech.
Lasecki \etal~\cite{Lasecki2012} employ non-experts to  collectively caption speech in real-time to help deaf and hard of hearing people.
Others~\cite{Zaidan2011} proposed a variety of mechanisms to 
collect high-quality translations for machine translation systems, 
and annotate text~\cite{Rosenthal2017,Nakov2016}.

In addition, crowdsourcing has been widely applied to acquisition of paraphrasing.
For example, Chklovski~\cite{Chklovski2005} designed a game to collect paraphrases with no prompting.
Negri~\etal~\cite{Negri2012} designed a set of paraphrasing jobs to maximizes the lexical divergence between an original sentence and its valid paraphrases.
Buzek~\etal~\cite{Buzek2010} proposed the idea of error-driven paraphrasing for machine translation systems. That is, they asked crowd workers to paraphrase only the parts of the input text that are problematic to the translation system. 
Burrows~\etal~\cite{Burrows2013} focused on the acquisition of passage-level samples using crowdsourcing while Lasecki \etal~\cite{Lasecki2013} collected dialog dataset.
Recently, Jiang \etal~\cite{Jiang2017} studied the key factors in crowdsourcing
paraphrase collection.

The design of \name{} has been inspired by many of the above approaches. However, most previous work cannot be readily applied to acquire adversarial dataset in natural language in an iterative manner.



\subsection{The effects of prompt}
When performing a task, crowd workers are influenced by instructions, examples, and context of the task~\cite{Jiang2017}. 
Some research focuses on how different prompts can result in natural variation of human-generated language.
For example, 
Wang~\etal~\cite{Wang2012} investigated three text-based elicitation methods, \eg, sentences, scenarios, or list-based descriptions, for collecting language that corresponds to a given semantic form.  
Mitchell~\etal~\cite{Mitchell2014} explored the use of crowdsourcing to generate a corpus of natural language templates for a spoken dialog system. They investigated the effect of presenting various amount of dialog content to crowd workers.
Kumaran~\etal~\cite{Kumaran2014} explored gaming as a strategy for acquisition of paraphrase data. This work presents drawing as prompt and asks the participants to produce paraphrases.
Law~\etal~\cite{Law2016} examined how crowd workers are incentivized by curiosity.  
In this work, we investigate how prompt can be augmented by machine learning to help crowd workers generate adversarial samples. 

\section{Formative Study}
\label{sec:formative}

The goal of the formative study is to understand current practice of model testing, the challenges faced by AI developers, and potential opportunities of our system.

\subsection{Study setup}
In this study, we interviewed five AI developers (denoted as D1---D5) in an IT company who are experienced in sentiment analysis. 
D1 is an engineer who has built sentiment classification models for different languages, such as German, English, and French.
D2 is a product manager who has analyzed errors in French sentiment models.
D3, D4 and D5 are research scientists who have experience in AI model design and cross-model evaluation. 


We organized semi-structured interview sessions with each expert. 
Each interview lasted approximately $30$ minutes and covered a variety of topics, starting with a general question about their experience in sentiment analysis, followed by how they test models' performance and their observation. 
We also focused on the challenges they encounter and how they address them. 
The interviewer took notes during the interviews and recorded audios for post-interview analysis. 
Based on the interview results, we derived four requirements to guide the design of \name{}.

\subsection{R1: Error generation}
To continuously improve the performance of sentiment models, the AI developers repeat the process of \textit{``Build (refine) model --- Train model --- Test model''}, where the results of model testing guide the refinement and training of models.
In model testing, the AI developers (D1, D2, D3, D4) mainly rely on metrics such as the entire accuracy of the testing dataset, the accuracy for each sentiment category, confusion matrix, F1 score, etc. 
One AI developer (D3) noted, \textit{``We built sentiment analysis models for research purpose, and evaluated our models by comparing with baseline approaches on an open dataset.''}
However, he was not sure about the performance of their model in real-world deployment, \textit{``If I need to deploy our model for real use, current testing would not be enough.''} 
Since existing testing dataset is limited in coverage, D3 suggested to borrow external dataset for a comprehensive testing, \textit{``the intuition is that, you need to increase the diversity of the testing data so as to cover different cases.''}
This motivates us to employ crowd force for error generation to extend the coverage of testing dataset.
In addition, we should allow AI developers to collect corpus of certain category to thoroughly test the performance of models, in particular regarding corner cases. 





\subsection{R2: Error validation}
After samples are crafted by crowd force, a critical task is to decide what are their ``real'' sentiment and whether the model makes correct predictions.
High quality testing dataset is critical for evaluating the performance of a model.
Some AI developers prefer human-labeled dataset because the quality is high.
That motivates us to borrow the crowd to manually validate the sentiment of each generated sample.
Since the sentiment is ambiguous and subjective, we plan to employ multiple crowd workers to validate one sample and use ``majority vote'' to mark as the ground truth.



\subsection{R3: Error categorization}
AI developers sometimes seek to obtain samples belonging to certain category to cover corner cases. 
For example, D2 mentioned that, ``We once tested the model for biasing. We tried Asian name and western people's names to see whether the model would give different predictions. We also tried like female names, male name to see if there any difference.''
Therefore, after obtaining samples generated by the crowd, it is critical to validate the category of them.
To deal with large sample size, labeling the category using the crowd is necessary to scale the labeling process.


\subsection{R4: Error analysis}
The analysis of mis-classified samples would reveal insights to the model. However, not all samples are worth the analysis.
As D2 mentioned, \textit{``If a model makes some error predictions, they may have different impact. For example, when a sentence is negative, but the prediction is positive, that would be polarity errors. People would think the model sucks. But for sentences that are ambiguous, for example, if the ground truth is positive, the model prediction is neutral, then it would be fine. Because people can understand these errors exist.''}
Therefore identifying mis-classified samples with high impact would help AI developers focus on the most important errors.
In addition, it would be infeasible to analyze all samples due to large sample size.
As a result, demonstrating the samples at multiple levels of granularities is necessary to deal with large volume of data.

\section{ChALLENGE.AI}
\label{sec:system}

\begin{figure}[tb]
    \centering
    \includegraphics[width=\linewidth]{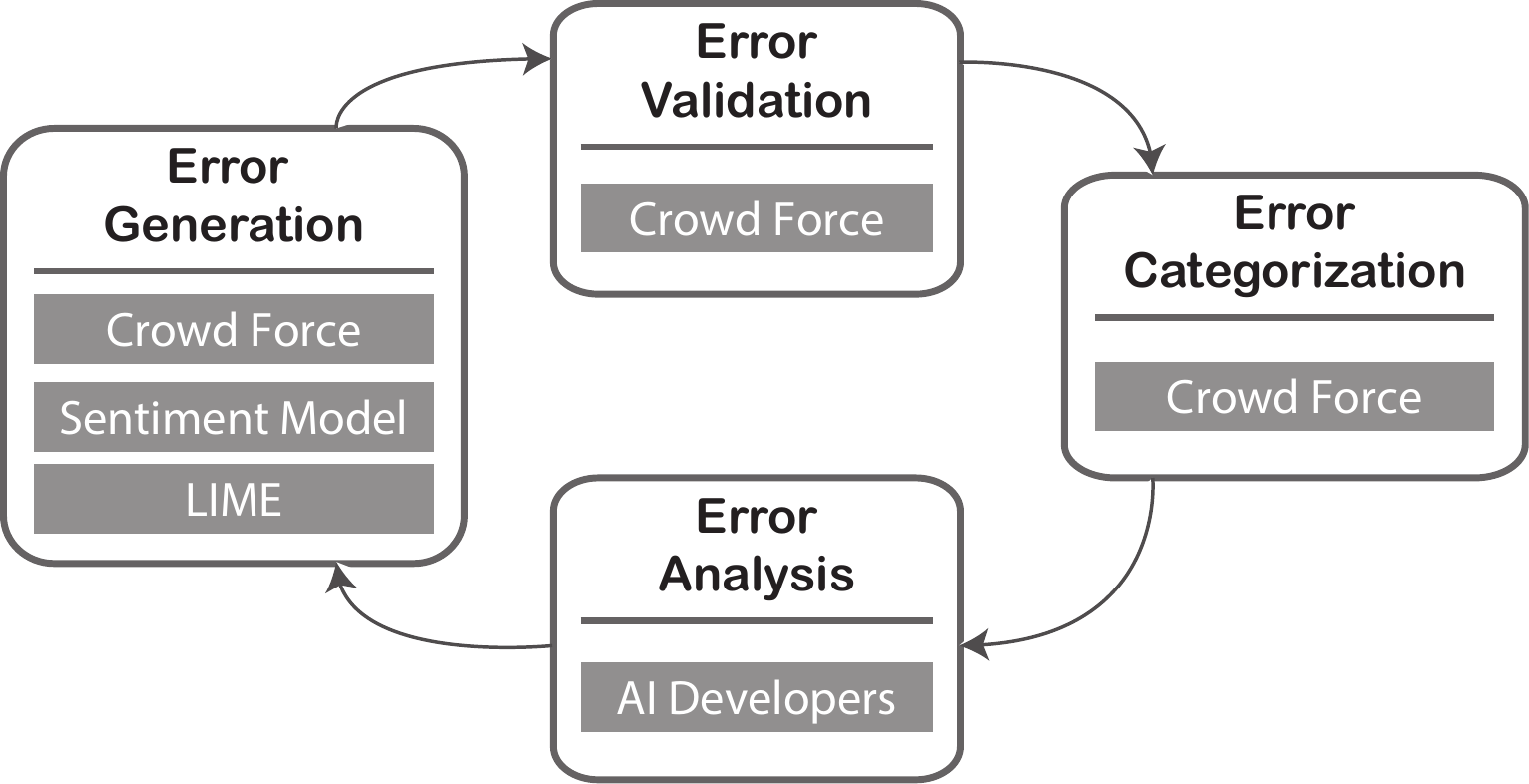}
    \caption{
        \name{} is an embodiment of proactive testing.
    This figure shows the architecture of \name{}. Our system starts from error generation, followed by error validation, categorization, and analysis by AI developers.
    Then AI developers can start over to iteratively test AI models using \name{}.
    }
    \label{fig:procedure}
\end{figure}

\name{} is a crowd system that assists AI developers in proactive testing.
Figure~\ref{fig:procedure} depicts the architecture of \name{} which includes four main components, \ie, explanation-based error generation, error validation, categorization, and analysis. 

\subsection{Explanation-based error generation}
This component is designed to encourage the crowd to craft sentences to fail AI models for evaluating the performance of the models.  
Recent approaches~\cite{Gao2018,Zhao2017,Cheng2018} based on deep learning have targeted on generating adversarial samples for text classifiers. However, these approaches cannot easily be extended to generate sentences of a certain category.
Additionally, their results do not guarantee to be legitimate from language viewpoint.
In this paper, \name{} combines the crowd intelligence  and machine learning to craft errors of any type.

\begin{figure}[!t]
    \centering
    \includegraphics[width=\linewidth]{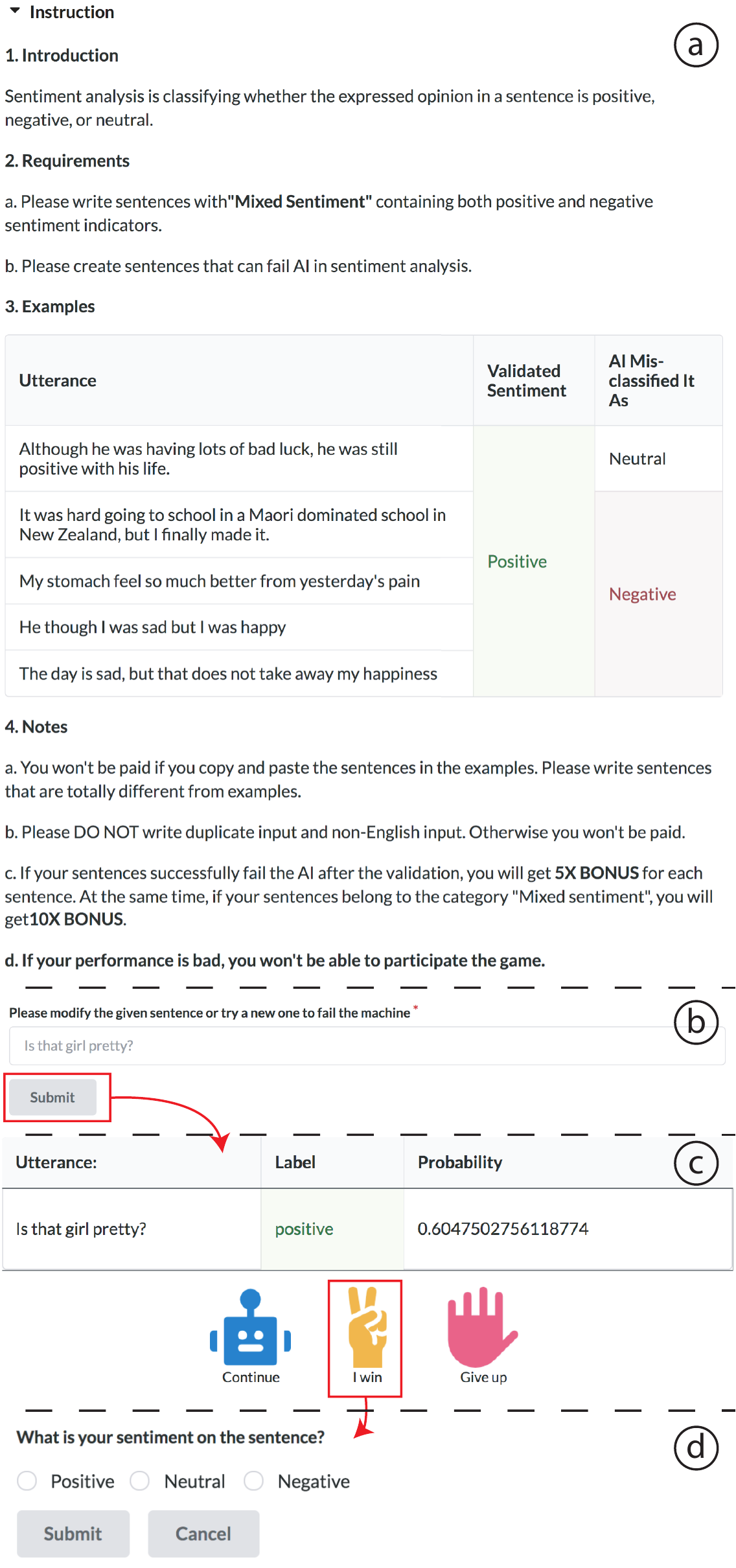}
    \caption{The excerpt of the Error Generation component.  (a) displays the instruction of the task. (b) When the interface is loaded, it shows an input area and a ``submit'' button below the instruction. (c) After a worker submits a sentence, \ie, ``Is that girl pretty?'', the interface displays the results  of sentiment analysis including a sentiment label and the probability. (d) After a worker clicks the ``I win'' button, a follow-up question is displayed. }
    \label{fig:generation}
\end{figure}

When the crowd enter the error generation component, the interface shows 
the introduction, example sentences belonging to a certain category, and rules of this task (Figure~\ref{fig:generation}(a)). 
After reading the instruction, a worker is able to craft a sentence in the input area. 
To avoid extremely short text, our system rejects input that is less than five words.
The worker then presses the ``Submit'' button to test the performance of the model (Figure~\ref{fig:generation}(b)). 
In response, \name{} launches the sentiment analysis model in the backend, and displays the analytics results, \eg, sentiment label (negative, neutral, or positive) and the probability in the result panel (Figure~\ref{fig:generation}(c)).
The worker can verify whether the model fails or not. If it fails, the worker then needs to identify the sentiment label of the sentence (Figure~\ref{fig:generation}(d)). Otherwise, the worker can decide to continue the trials or give up.



To encourage crowd workers to generate text that meets our requirements, we offer ``effort-responsive'' bonus~\cite{Gaikwad2016}. For example, a worker can get five times of bonus if the proposed text really fails the analyzer. Further, if the text belongs to the target category, ten times of bonus is offered to the worker.

\subsection{Combining crowd force and machine learning}

\begin{figure*}[tb]
    \centering
    \includegraphics[width=\linewidth]{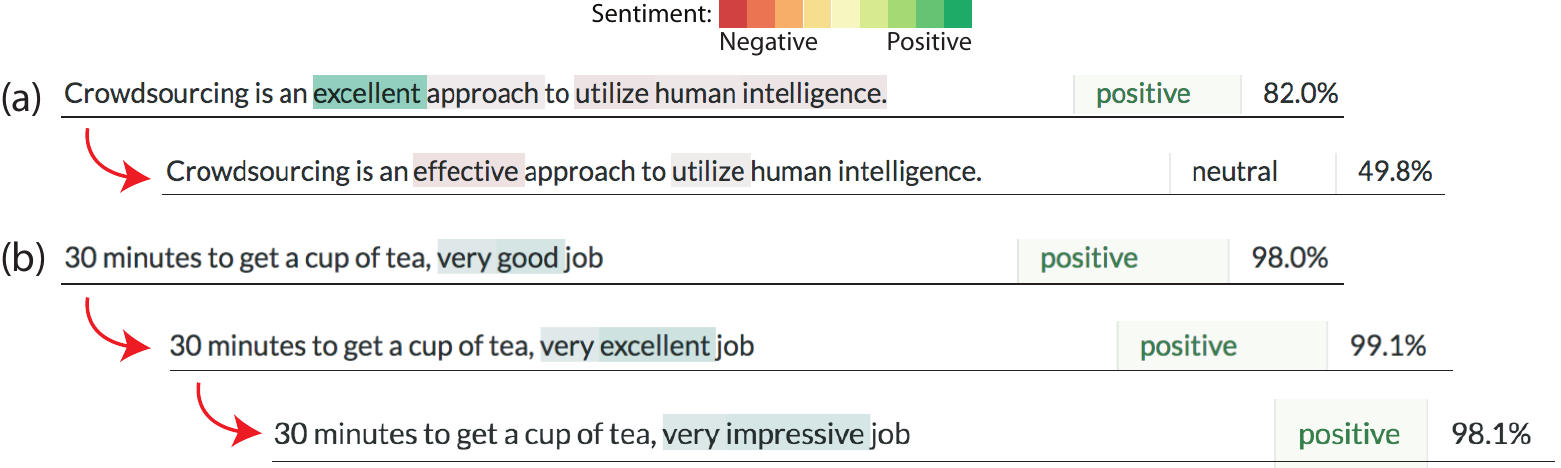}
    \caption{The usage of LIME in two cases. (a) shows how LIME helps crowd workers modify the input sentence to successfully fool the analyzer. (b) demonstrates how LIME facilitates workers to continuously generate adversarial samples.}
    \label{fig:lime}
\end{figure*}

Surrounding context may have an effect in facilitating crowd workers to craft samples, affecting the performance such as efficiency, quality, and success rate, etc~\cite{Jiang2017}.
We seek to augment the prompts to assist crowd workers in error generation from two aspects, \ie, starting point and accountability, respectively. 
The starting point refers to the existing text in the input box (Figure~\ref{fig:generation}(b)), we boost the crafted sentences by providing a randomly sampled error from one category.
Crowd workers are encouraged to edit the sentence in the input area.

On the other hand, 
we provide accountability by borrowing LIME~\cite{Ribeiro2016}, an explanation technique that provides interpretable results for a prediction and is applicable to explain any models.
To be specific, after a worker submits a sentence, the LIME algorithm is triggered to calculate the relationship between the prediction and each word in real time.
Then the results are presented in the interface.
Instead of presenting a set of numeric values, we borrow visualization techniques to intuitively depict the LIME results inline with the text. 
As shown in Figure~\ref{fig:lime}(a), 
The background color  of a word indicates whether it contributes to positive (green), negative (red), or neutral (yellow) sentiment.
We apply a diverging color scheme from ColorBrewer~\cite{Harrower2003} to indicate the polarity of the sentiment. 



Here are two working scenarios describing how the LIME results assist the crowd to revise the input for the next iteration.
Jane first crafted a positive sentence, ``Crowdsourcing is an excellent approach to utilize human intelligence.'', and found that the analyzer is able to recognize the sentiment. 
The LIME result (Figure~\ref{fig:lime}(a))  shows that the word ``excellent'' has a green background, 
which indicates that  ``excellent'' contributes a lot to the positive sentiment. 
Therefore, in the next iteration, Jane changed ``excellent'' to ``effective'' to lower the tone. 
And she successfully fools the analyzer to recognize it as neutral, as shown in Figure~\ref{fig:lime}(a).
Jane further generates a negative sentence using sarcasm, ``30 minutes to get a cup of tea, very good job''. 
The sentence successfully fools the model to recognize it as positive, and the phrase ``very good'' contributes a lot to the prediction. 
Then Jane modifies the sentence by changing the word ``good'' to other positive words such as ``excellent'' and ``impressive'' to keep fooling the model.
In Section~\ref{sec:eval1}, we quantitatively evaluate how LIME assists crowd workers in adversarial error generation.

\subsection{Error validation}
We conduct crowd-based validation by recruiting different crowd workers after the Error Generation process to obtain ground truth sentiment labels, such as positive, neutral, and negative. 
In addition, we offer ``effort-responsive'' bonus to creators based on the validation results.

During generation, each worker can craft an arbitrary number of sentences until they give up, and mark whether the crafted sentences have failed the model (by clicking the ``I win'' button) or not (by clicking the ``Continue'' button in Figure~\ref{fig:generation}(c)). 
We only validate sentences that are labeled to have failed the model. 

\begin{figure}[tb]
    \centering
    \includegraphics[width=\linewidth]{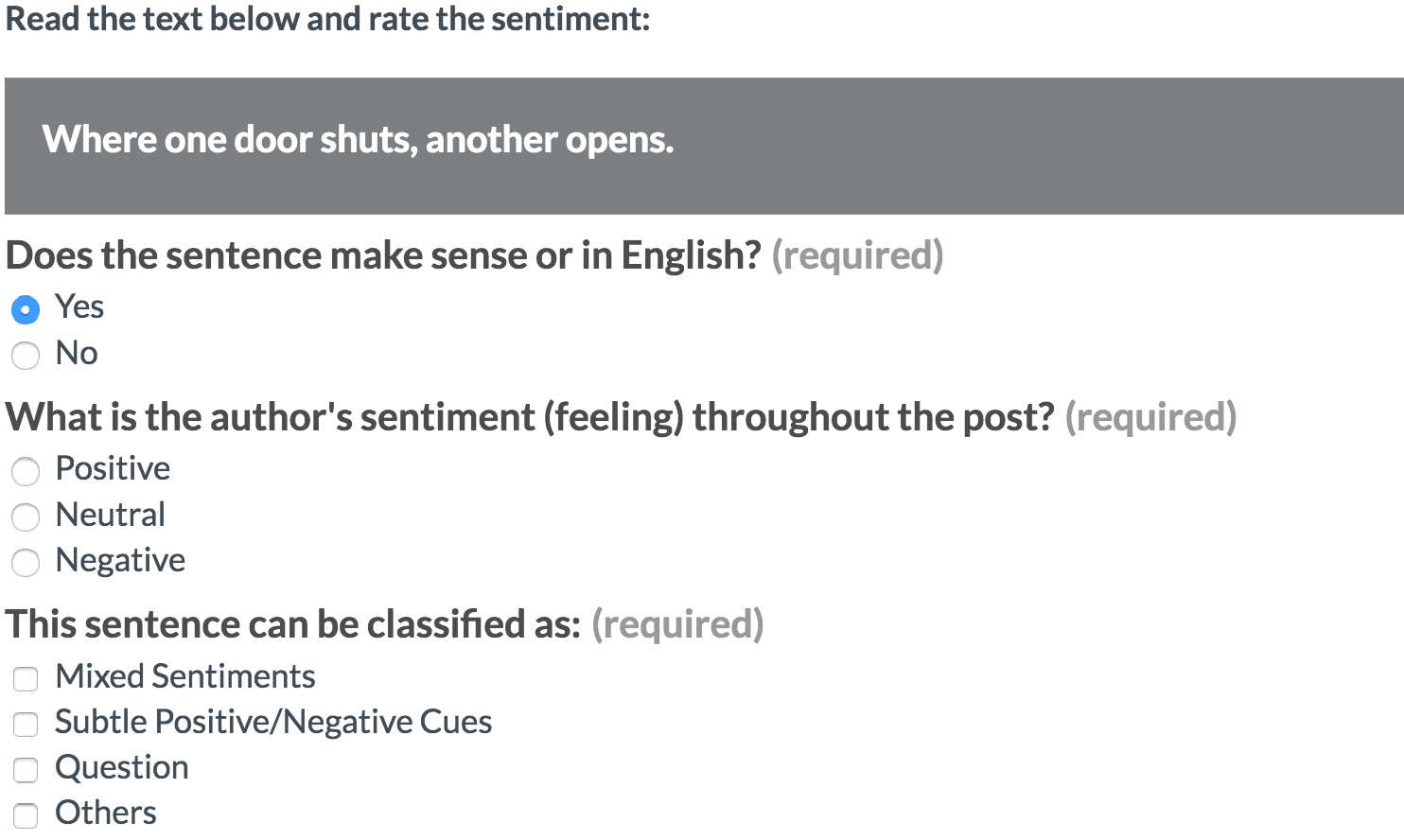}
    \caption{The excerpt of the validation interface on a crowdsourcing platform.}
    \label{fig:validation}
\end{figure}

To ensure the quality of these samples, we propose two criteria that a good sample should minimally satisfy. 
First, the sample should be in English and syntactically correct from the language viewpoint.
Second, it should successfully fail the analyzer.
Figure~\ref{fig:validation}(a) displays the excerpt of the validation interface. 
Crowd workers are asked to verify whether the sentence is in English and makes sense. If so, they continue to finish the remaining questions for sentiment validation. 

We require at least $5$ judgments for each sample, and pay $\$0.016$ per judgment.
We set up many hidden test questions for quality control, which are used to reject validations by workers who have missed a quantity of test questions~\cite{Rosenthal2017}.  
The validation is performed using Figure-eight~\cite{f8}

\subsection{Error categorization}
To test whether crafted sentences belong to the target category,
we employ the crowd to perform categorization for each sample.
In practice, categorization is preformed along with validation as shown in Figure~\ref{fig:validation}(b). 
As described in the Section \textbf{Evaluation with AI developers}, initial categorization is identified by AI developers, and the categories are iteratively refined during the \textbf{Error Analysis} step.




\subsection{Error analysis}

\begin{figure*}[tb]
    \centering
    \includegraphics[width=\linewidth]{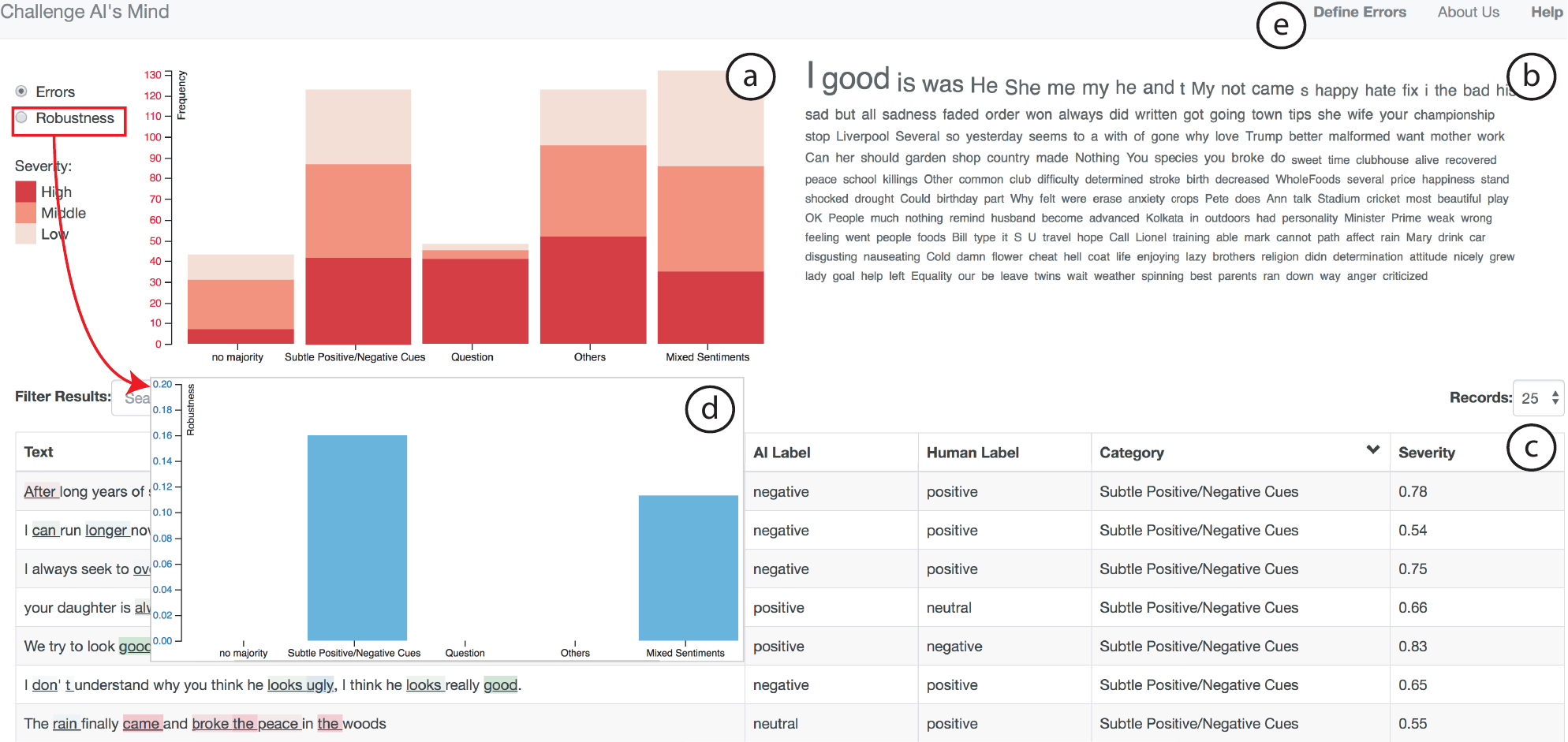}
    \caption{The interface allows AI developers to investigate the validated samples at different levels.}
    \label{fig:expertUI}
\end{figure*}

After error categorization, we obtain a dataset where each error is associated with a ground truth sentiment label validated by the crowd, a predicted label by the model, and a category. 
To understand the impact of each error, we define ``Severity'' for each error. 
The intuition is that, for a misclassified sentence, if both human and the model are confident about the sentiment, the mistake is severe. 
On the other hand, if both sides are not sure about the sentiment label, the mistake can be ignored.
Hence, the severity score is calculated as:
\[ 
S=W_1 \times \text{Conf}_{human}+W_2 \times \text{Conf}_{AI}
\]
Where $W_1$ and $W_2$ are weights for confidence of human and of the model, respectively. In this work, we set $W_1=W_2=1/2$. 
$\text{Conf}_{human}$ represents the confidence of human, which is calculated as the percentage of the crowd making the judgment the same as majority vote. For example, for a sentence validated by five crowd workers, three of them validated the sentiment as positive. Hence, $\text{Conf}_{human}$ becomes 
$3/5=0.6$.
Where $3$ is the number of workers making the judgment as positive (majority vote) and $5$ is the total number of workers.
$\text{Conf}_{AI}$ is provided by the model, usually obtained as the probability or confidence of the prediction.

To help AI developers understand the model by analyzing a large quantity of errors, we build an interface to demonstrate the analysis at three different levels.
After the data is loaded, the Statistic View (Figure~\ref{fig:expertUI}(a)) uses a stacked bar chart to demonstrate the error distribution of each category at the macro-level.
The x-axis presents different categories while the y-axis shows the number of errors.
For each category, we manually set two thresholds to split each category into three rectangles representing the number of samples with high severity (dark red), middle (light red), and low (pink).
In addition, to demonstrate how easy the crowd craft samples for a target category, we borrows a bar chart (Figure~\ref{fig:expertUI}(d)) to display the ``Robustness'', which is defined as,
$$
\text{Robustness}=\frac{N_{categ}}{N_{valid}}
$$
Where $N_{valid}$ means the number of sentences that successfully fail the model based on the validation results, and $N_{categ}$ is the number of sentences that can both fail the model and belong to the target category.
Figure~\ref{fig:expertUI}(d) shows that the robustness of `Subtle positive/negative cues' is about 16\% which is higher than `mixed sentiment'. That means the crowd has higher success rate in making sentences belonging to `Subtle positive/negative cues'.
At the meso-level, a Cloud View (Figure~\ref{fig:expertUI}(b)) shows a tag cloud summarizing sentiment words calculated by LIME~\cite{Ribeiro2016}.
The bigger a word is, the more frequent it appears in sentences as a sentiment word recognized by LIME.
At the micro-level, a Table View (Figure~\ref{fig:expertUI}(c)) demonstrates raw sentences, the prediction, sentiment ground truth, the category, and the severity. 
Each sentiment highlights sentiment words using underline and different background color.
Red indicates negative words while green is positive and blue is neutral.

Various interaction techniques, such as linking and filtering, are borrowed to coordinate the three views.
For example, 
when clicking on a bar in the Statistic View, the tag cloud in the Cloud View will update to show sentiment words of this category.
When a sentiment word is clicked in the Cloud View, raw sentences in the Table View will be filtered to display all sentences including that word.
In addition, the interface allows AI developers to define new sample categories (Figure~\ref{fig:expertUI}(e)) and start over the proactive testing pipeline from error generation.



\section{Evaluation with the Crowd}
\label{sec:eval1}
We conducted a crowd evaluation to investigate how different prompts in error generation affect the performance of the crowd in crafting errors.


\subsection{Methodology and Procedure}
We constructed prompts based on different combination of accountability (LIME) and starting points (SP). 
Accountability refers to the explanation of a prediction using LIME~\cite{Ribeiro2016}. The two options are displaying the results of LIME to the crowd (\textbf{LIME}), or not (\fade{LIME}).
According to our interview with five AI developers (as discussed in Section~\ref{sec:eval2}), 
A starting point can be either empty (\fade{SP}), or a randomly sampled misclassified sentence from one category.
If the starting point is empty, workers are encouraged to craft a sentence from scratch.
Otherwise, workers are allowed to edit the text in the input area (\textbf{SP}). 
We identified two types of errors to generate, \ie, ``Mixed-sentiment'' and ``Subtle sentiment cues'', based on the results of the first sessions described in Section \textbf{Evaluation with AI Developers}.

We performed a between-subject design with two experimental conditions, \eg, a baseline condition (\fade{LIME}, \fade{SP}) and an enhanced prompt (\textbf{LIME}, \textbf{SP}), and two types of errors for error generation.
We used Figure-eight~\cite{f8} as the platform to release our error generation jobs, 
and created pilot runs to help us refine the instruction and test \name{} in the wild.
To match the going rates on Figure-eight, we paid \$0.01 per participant. 
To be slightly generous, we paid \$0.05 per sentence if the sentences successfully fail the model after validation. 
At the same time, if the sentences belong to the required category, additional \$0.05 per sentence were paid to the crowd.
To reject noises and assign categories to each sample, the crowd-based validation was performed after generation. 

\subsection{Metrics}

\begin{table}[ht]
	\centering
	\begin{adjustbox}{width=.85\linewidth}
		\small
		\begin{tabular}{lccc}
			\toprule
                  & {\textbf{LIME} \textbf{SP}} & {\fade{LIME} \fade{SP}} & Total \\
			\midrule
			
			$N_{total}$ & 262                         & 293                     & 555   \\
			$N_{valid}$ & 75                          & 108                     & 183   \\
			\# workers  & 66                          & 46                      & 112   \\
			
			\bottomrule
		\end{tabular}
	\end{adjustbox}
	\label{tab:generalStat}
	\caption{Statistics of error generation based on two prompt conditions, \eg, a baseline condition (\fade{LIME} \fade{SP}) and an enhanced one (\textbf{LIME} \textbf{SP}).
	}
	\label{tab:generalStat}
\end{table} 

The general statistics of each job are displayed in Table~\ref{tab:generalStat}. 
The \textbf{Total trials}, denoted as $N_{total}$, include all sentences that the crowd have crafted using our system.
Crowd workers have generated $249$ sentences for ``Subtle sentiment cues'' and $306$ for ``Mixed-sentiment'', respectively.
Since we do not limit the number of sentences the crowd craft in the generation process, the number of trials varies across different conditions.
To obtain the ground truth of the sentiment label for each sentence, we validate sentences marked to have failed the model during the validation process.
\textbf{Validated trials} ($N_{valid}$) are number of sentences that successfully fail the model based on the ground truth.
In addition, we count the number of distinct crowd workers for each condition (\textbf{\# workers}).
Accordingly, we propose three metrics to evaluate the performance of each crowd worker. We use $n$ instead of $N$ to represent that the statistics values correspond to one crowd worker.

\textbf{Average time per trial ($T$)} measures how much time that a worker needs to craft a trial on average.
We assume that each trial does not exceed five minutes.
Therefore, for trials made by one worker, the time of each trial is calculated by 
$T_{trial}= min(300~seconds, T_{current}-T_{previous})$.  
Then we obtain: $T=\sum_{i=1}^{n}{T_i}/n$, where $n$ is the number of all trials made by one worker.
$T$ measures how efficient a sentence can be crafted. 

\textbf{Success rate ($R_{succ}$)} is measured as 
$n_{valid}/n_{total}$. 
This value measures how easily a worker thinks s/he can generate samples to fail the model. 
The success rate is useful to measure the effectiveness of prompts, as well as to analyze the vulnerability of a model.

\subsection{Analysis of crowd performance}
\begin{figure}[tb]
	\centering
	\includegraphics[width=\linewidth]{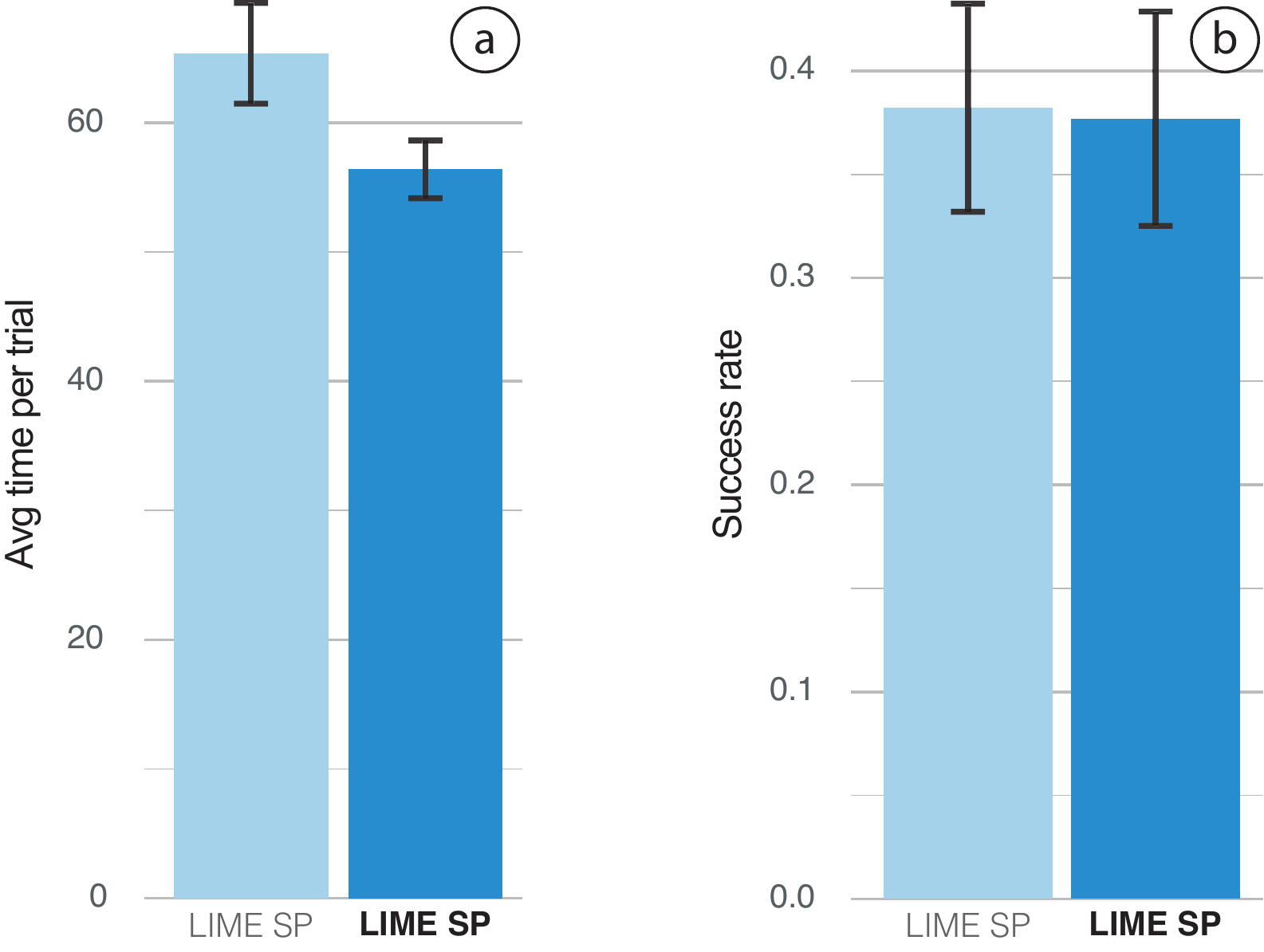}
	\caption{(a) shows the bar chart displaying average time per trial for each worker under two conditions. (b) shows how crowd workers differ in success rate. The error bars demonstrate standard errors.
	}
	\label{fig:bar}
\end{figure}

On average, crowd workers spent 56.4 seconds (SD=18.2) in crafting a sentence with the enhanced prompt (\textbf{LIME} \textbf{SP}) and 65.4 seconds (SD=26.4) with (\fade{LIME} \fade{SP}). Figure~\ref{fig:bar}(a) shows the average time per trial ($T$) under each condition.
We found a significant effect (t = 1.9977, p<0.05) of the enhanced prompt in reducing ($T$). The crowd used about 13.8\% less time in crafting a sentence with (\textbf{LIME} \textbf{SP}) than (\fade{LIME} \fade{SP}).
The reason may be that editing text in the input area requires less time compared to crafting a new sentence from scratch.

Figure~\ref{fig:bar}(b) shows that crowd workers are indifferent in success rate ($R_{succ}$) under two conditions (38.2\% V.S. 37.7\%). 
This result might be because crowd workers are not familiar with the color encoding of LIME explanation (Figure~\ref{fig:lime}) and randomly sampled sentences do not help the crowd to craft sentences to fail the model.

Besides qualitative assessment, we received positive feedback from the crowd regarding the error generation tasks.
For example, one of them commented, 
\textit{``This was fun, I sure hope my answers were good. If not please dont pay me, I enjoyed the task and want to be able to try some more in the future.''}

To conclude, the enhanced prompts help the crowd use less time in crafting a sentence that can fail the model. To understand how the two key factors, \eg, accountability and starting point, interact with each other,  we plan to recruit more crowd workers for error generation and perform a two-way ANOVA for detailed analysis in future research. 

\section{Evaluation with AI developers}
\label{sec:eval2}

To investigate how \name{} helps AI developers understand and diagnose a model,
we worked with the five AI developers that we collaborated during the formative study,
and organized two rounds of semi-structured interview sessions to evaluate the 
effectiveness and usefulness of \name{}.

\subsection{Process}
We followed the architecture of \name{} (Figure~\ref{fig:procedure}) to evaluate the entire system.
Before error generation, we started from the first sessions with AI developers to obtain initial categorization for errors.
Based on the category information proposed by AI developers, we used \name{} to generate errors belonging to these categories, and conducted validation and categorization for crafted sentences.
Finally, we organized the second interview sessions (error analysis) to understand the usefulness and limitations of \name{} from the perspective of AI developers.
During the entire evaluation, we used a sentiment analysis model built by D1 as the target model to test. 
The input of the model is a sentence, and it outputs a sentiment label associated with a probability.

\subsubsection{First sessions}
The goal of the first sessions is to 
obtain the target categories of errors to test the model.
To begin with, we tested the performance of the model using a public sentiment dataset~\cite{Rosenthal2017} where all \siwei{12284} sentences are collected from Twitter, and labeled with negative, neutral, or positive sentiment.
After obtaining all misclassified sentences, we randomly sampled $200$ ones and 
stored them in a table (CSV file format) with four columns, \eg, a `Text' column, 
a `Human\_Label' column showing the ground truth, 
an `AI\_Label' column displaying the results calculated by the model, 
and an empty column titled `Category' to allow AI developers to label a potential category for the sentence.

Each interview started with the introduction of the dataset. 
After that, we presented the dataset to AI developers and
asked them to identify the patterns of the misclassified samples and name new categories for them.
AI developers were allowed to discard sentences that are hard to be categorized.
An interview took about $40$ minutes.
We encouraged them to express findings and thoughts using a think-aloud protocol and took notes about their feedback for further analysis.

Some AI developers have more experience in identifying patterns for errors.
For example, when noticing a sentence whose benchmark label is positive, but misclassified as negative by the model, \ie,
``Marissa Miller of Google makes shout out to the  Khan Academy and the great things they're doing for education. \#fmsignal \#sxsw (cc @mention'', 
D2 said, \textit{``I think the model made a wrong prediction because it does not understand what `shout out' means.''}
From her experience, D2 further commented that the model may not understand sentiment indications that are domain-specific or context dependent.
Besides summarizing patterns in the dataset, D3 asked for sentences containing both positive and negative indicators.
``Do any of them have opposite sentiment words, like, I am happy, but... something like that?''
The participant further explained, \textit{``Some models are designed to handle targeted sentiment, but determining relevant sentiment in mixed sentiment texts is challenging.''}
Finally we derived two categories of errors for model testing. One is called ``Subtle Sentiment Cues'' which means that a sentence is either positive or negative, and has positive or negative indications.
The other is ``Mixed-sentiment'' which refers to sentences containing both positive cues and negative indicators. 
Further, we include three more types of errors for categorization.
For example, a ``Questions'' category is added based on D1's comments and an ``Others'' is included to be more general. 
A ``No majority'' category is added after categorization if human annotators cannot reach a consensus on the category of that sample.


\subsubsection{Running \name{}}
After obtaining the categorization, we tested the model by walking through three main components of \name{}, \eg, error generation, validation, and categorization.
As mentioned above, we focused on the two categories, \ie, ``Subtle Sentiment Cues'' and ``Mixed-sentiment'' in error generation while we used five categories for error categorization.
The results and analysis of crowd performance are described in Section~\ref{sec:eval1}.

\subsubsection{Second interview sessions}
We organized second interview sessions to evaluate how \name{} helps AI developers understand the performance of the model.

After running \name{}, we obtained $555$ samples that $112$ crowd workers generated to have successfully failed the model, where $23$ errors are categorized as ``Subtle Sentiment Cues'' and $44$ are ``Mixed-sentiment''.
During the interviews, we demonstrated the data at three levels of granularities using the interface shown in Figure~\ref{fig:expertUI}.

Each interview took about $45$ minutes. 
We first presented the goal of \name{} to AI developers and a detailed introduction to the data and interface.
AI developers then freely explored the interface and we helped them resolve any questions they encountered. 
Next, the participants went through the interface to tell how they understood the performance of the model.
They further identified new categories of errors by investigating detailed samples using the interface.
Finally, a post-interview discussion was conducted to collect their feedback about the strengths and weaknesses of \name{}.
During the interview, AI developers were instructed to think aloud and we took notes about their feedback. We recorded the whole interview sessions for later analysis.
We report the results of second interview sessions in the remaining of the section.

\subsection{Value of proactive testing}
A thorough testing is important for AI models before deployment. 
However, current practice of testing is limited in coverage, 
as D3 commented, \textit{``When doing the testing, we assume that the testing dataset and training dataset are in the same feature space.''}
Traditional testing approach is far from enough for deploying the model in the wild, which
indicates the potential value of proactive testing in evaluating the model for production. 
To reduce critical and embarrassing errors, AI developers are able to identify corner cases to test, and \name{} collect external dataset belonging to specific categories.
In addition, by investigating external dataset, AI developers can discover unseen errors.
For example, our participants identified two categories that are distinct from those found in the first interview session, \eg, bias in pronouns such as `He' and `She', and reversed sentiment containing words like `However', `Though', and `But'.
Detailed discussions are reported below.



\subsection{Getting a gist}

First of all, AI developers were interested in the overall patterns of misclassified samples.
The Statistics View (Figure~\ref{fig:expertUI}(a)) provides a big picture of the entire dataset. 
From the stacked bar chart, D5 noticed that it is about equal distribution among high severity, middle, and low for most bars. However, the samples belonging to ``Question'' attracted her attention because high-severity errors account for the majority in this category.
\textit{``The model could be improved (in the `Question' category) for sure.''}
D5 further explained the way of improving the model,  \textit{``In some of the supervised learning models, we need to use human heuristics to do the feature engineering (extraction) from the raw dataset. The quality of the feature extracted largely impacts the final performance.''}
The participant took the ``Question'' category as an example, \textit{``If a model a has high probability to make severe errors for question sentences, we may specify a feature in feature engineering to detect whether a sentence is a question or a statement. So with this feature, hopefully could help the model make decisions.''}

From our observation of the first sessions, all AI developers had read through about a dozen of misclassified sentences because the process of error analysis requires great mental efforts. 
Displaying the errors at different levels of granularities would relieve AI developers in analyzing a large number of errors.
As D2 commented, \textit{``I like the overview which gives me the impression of the entire dataset. You know, reading through two hundred errors is time-consuming and impossible (during the first interview session), and I did not do a good job last time.''}



\subsection{Examining errors by words}
After examining the Statistics View, D4 switched his focus to the Cloud View showing sentiment words as tag cloud (Figure~\ref{fig:expertUI}(b)). The participant noticed that the word ``I'' has the biggest font size while ``Good'' is the second biggest word. \textit{``Typically in sentiment analysis, you will not expect `I' to be particularly positive or negative. `Good' is the second one.  It makes more sense but `I', `is', `was', `he', `me', `my', `she', among the first line are not sentiment words.''} 
However, the participant changed his mind after investigating sentences containing ``He'' and ``She''.
He first clicked ``She'' and the Table View updated. The participant noticed that the word contributes a lot to neutral sentences, and contributes once for negative and positive, respectively.
Similarly, the participant further examined sentences containing the word ``He'', and noticed that four out of eight are negative, and ``He'' contributes to the negative sentiment. 
\textit{``Well, it is interesting to see the difference between `She' and `He'. I guess the model tends to regard `He' as a negative word.''} 
He added, \textit{``I think that it is necessary to examine the training data (of the model) to see whether the stop words are equal in distribution for each sentiment.''}

Before using \name{}, some AI developers (D1, D4, and D5) found it hard to identify patterns and categorize sentences. For example, during the first interview sessions, D4 did not know the reason for some of the predictions. The participant pointed to one question sentence and commented,
\textit{``There is no reason to label this question into negative or positive. Because it apparently contains none of the words with any sentiment.''}
D4 and D5 noted that they did not agree with some ground truth labels. As D4 said, 
\textit{``I would recommend you have a category for mis-labeled because it is subjective.''} The participant further pointed to a sentence whose benchmark label is neutral, and added, 
\textit{``Now here is one, `Social Is Too Important For Google To Screw Up A Big Launch Circus'. 
It sounds kind of negative to me, which is how the model classified it as.''}
By borrowing LIME~\cite{Ribeiro2016} to extract sentiment words, \name{} provides explanation of errors at the word level, allowing AI developers to find potential bias in the training data.

\subsection{Reading through errors}
D1 showed great interest in the exploration of samples in the ``Mixed-sentiment'' category. 
He clicked bars with dark red color under this category and read through these severe errors in the Table View.
Then the participant noted, 
\textit{``Some sentences in this category are reversed sentiment.''} Then the participant pointed to a sample and added, \textit{``Like in this case, it has the word `but'. All content after `but' is the content that the speaker wants to emphasize. The former part is like warm up. So the later part highlights the whole meaning of the sentence. In this case, I will not say it is a mixed sentiment. It is reversed.''}
Then, the participant used the search box to find all sentences containing ``however'' but found no sample in the table. 
He commented, \textit{``I would like to test the model with sentences using reversing words, like `but', `however', `although', etc. The model may not do a good job.''}

During the first interview sessions, we realize that not all errors are worth investigation.
When looking at the errors, D5 commented, \textit{``A lot of these are difficult for human. For those which are less obvious, you may ask three different people and got three difficult answers.''} The participant further added, \textit{``Since sentiment analysis is subjective, if an error is ambiguous to human, I do not think the model made a severe mistake.''}
Therefore, the definition of severity helps AI developers focus on errors that are important to examine.

\section{Design Implications}


Proactive testing is a promising direction that helps AI developers get more insights into the model. 
\name{} is the first prototype that supports proactive testing using the crowd force, and
we suggest the following aspects that future research can explore.

First, \textit{include all the generated data by the crowd including those that can fail the model and those cannot.}
Because only the misclassified samples are not enough to help AI developers understand how the model performs in some cases.
For example, D2 has found two sentences containing the word ``Trump'' by filtering. 
However, the participant could not conclude whether the model is biased to the word ``Trump''. 
D2 commented,
\textit{``I am only looking at the errors. It is hard to tell (whether the model is biased to ``Trump'').
I mean, these errors could be 99\% of the instances in which case the model is doing very poorly. But this could be less than 1\% of the instances in which case the model is doing fantastic.''}

Second, \textit{apply better explanation techniques.} 
In this study, we choose the LIME algorithm~\cite{Ribeiro2016} to identify and highlight sentiment words related to the prediction. 
However, our participants found that some sentiment words are confusing.
For example, D4 found a positive sentence with AI labeled negative, ``I can run longer now''. The word ``can'' is highlighted in green (positive) and ``longer'' highlighted in blue (neutral).
He commented, 
\textit{``The AI label is negative. However, it is wired that no words are marked as negative.''}
However, when more advanced analytical techniques are developed in the future, such issue may be resolved.


Third, \textit{enhance the generation component for \textit{word-level} categories.}
\name{} has been proved to be effective in collecting samples belonging to \textit{concept-level} categories such as ``mixed-sentiment'' and ``subtle sentiment cues''. 
However, AI developers may sometimes seek to test the model using samples containing certain words, such as ``Trump''.
Intuitively, collecting samples with certain words could be more cost- and time-efficient by using techniques in information retrieval.
We plan to study how various information retrieval techniques help in collecting samples of different category.

Fourth, \textit{provide real-time feedback for proactive testing.}
The main process of sample collection, \eg, generation, validation, and categorization, takes a long time and AI developers cannot test the model in real-time. 
One possible solution is to borrow workflows from real-time crowdsourcing~\cite{Lasecki2011,Cruz2015,Lundgard2018,Liao2018} to reduce the delay in obtaining the testing results.
Another solution is to augment the error analysis interface as suggested by D2,
\textit{``Since the model is already trained. Maybe you can (embed the model in the backend and) add an input box for real-time testing so that I can test some of the sentences in my mind.''}

Fifth, \textit{augment error analysis with advanced analytical methods.}
our system borrows knowledge from AI developers to identify new patterns to test. 
However, the process is time-consuming and not scalable. 
It would be beneficial to incorporate automatic analytical methods, such as text classification or clustering, to assist AI developers in summarizing patterns among errors.

\section{Generalizability and Future Work}


Although our study grounds the exploration in the context of sentiment analysis, our system can be easily generalized to other text classification domains for crowd proactive testing, such as part-of-speech analysis. 
In addition, we found that explanation helps the crowd craft samples that fail the model.
This idea can be borrowed in the generation of adversary samples using crowd intelligence in other fields, such as computer vision, for adversary learning.

There are a number of promising future directions. 
First, after error categorization, we obtained a testing dataset where each sample is labeled with a ground truth category and sentiment. 
We plan to release the dataset to the public to benefit more AI developers in testing sentiment analysis models.
Second, we seek to have a comprehensive understanding of the crowd-crafted dataset by analyzing it from different perspectives.
We plan to establish metrics to compare the generated dataset and open sourced ones from different perspectives, such as the distribution of sentence length, topic coverage, syntactic structure, uni-gram distribution, etc.

\section{Conclusion}
In this paper, we designed and built a new crowd system \name{} for AI developers to  proactively test their models. \name{} consists of four components including error generation, validation, categorization, and analysis.
Our system features an explanation-based error generation component that incorporates crowd intelligence and machine learning to facilitate the crowd in crafting errors to fail a sentiment analysis model.
We conducted a crowd user study to quantitatively evaluate the effectiveness of the explanation-based error generation component, and we found that the explanation-based error generation technique saved crowd workers 13.8\% in crafting sentences to fail the model. We also evaluated \name{} with five AI developers and
the study showed that the system helped participants to identify new error categories that have not been discovered before. We believe the proactive testing architecture developed in this work offers new opportunities and tools to reshape AI testing process.

\balance{}

\bibliographystyle{SIGCHI-Reference-Format}
\bibliography{sample}

\end{document}